%% file: iclr2024_conference.tex
\documentclass{article} % For LaTeX2e
\usepackage{iclr2024_conference,times}

\input{math_commands.tex}

\usepackage{hyperref}
\usepackage{url}
\usepackage{graphicx}
\newcommand{\tabitem}{~~\llap{\tiny$\bullet$}~~}
\usepackage{placeins}
\usepackage{verbatim}
\usepackage{multirow}
\usepackage{subcaption}
\usepackage{amsthm}
\usepackage{booktabs}

\title{DON-LSTM: Multi-Resolution Learning with DeepONets and Long Short-Term Memory Neural Networks}

\author{Katarzyna Michałowska\thanks{Corresponding author: katarzyna.michalowska@sintef.no} 
\\
SINTEF AS \& University of Oslo\\
\And 
Somdatta Goswami \\
Brown University \hspace{37.5pt} \\
\AND
George Em Karniadakis \\
Brown University \\
\And 
\And 
\And 
\And  
\And 
\And 
\And 
Signe Riemer-Sørensen \\
SINTEF AS\\
}

\iclrfinalcopy % Uncomment for camera-ready version, but NOT for submission.
\begin{document}

\maketitle

\begin{abstract}
%The abstract paragraph should be indented 1/2~inch (3~picas) on both left and
%right-hand margins. Use 10~point type, with a vertical spacing of 11~points.
%The word \textsc{Abstract} must be centered, in small caps, and in point size 12. Two
%line spaces precede the abstract. The abstract must be limited to one
%paragraph.
Deep operator networks (DeepONets, DONs) offer a distinct advantage over traditional neural networks in their ability to be trained on multi-resolution data. This property becomes especially relevant in real-world scenarios where high-resolution measurements are difficult to obtain, while low-resolution data is more readily available. Nevertheless, DeepONets alone often struggle to capture and maintain dependencies over long sequences compared to other state-of-the-art algorithms.
We propose a novel architecture, named DON-LSTM, which extends the DeepONet with a long short-term memory network (LSTM). Combining these two architectures, we equip the network with explicit mechanisms to leverage multi-resolution data, as well as capture temporal dependencies in long sequences. We test our method on long-time-evolution modeling of multiple non-linear systems and show that the proposed multi-resolution DON-LSTM achieves significantly lower generalization error and requires fewer high-resolution samples compared to its vanilla counterparts. 
\end{abstract}

\section*{Introduction}

Modeling the temporal evolution of dynamical systems is of paramount importance across various
scientific and engineering domains, including physics, biology, climate science, and industrial predictive maintenance. Creating such models enables future predictions and process optimization, and provides insights into the underlying mechanisms governing these systems. Nevertheless, obtaining precise and comprehensive data for modeling poses a challenge in practical contexts. Real-world scenarios bring along the issue of data limitations, where data is typically more readily available at lower resolutions: historical measurements are stored with reduced granularity, vast quantities of legacy data originate from an outdated technology, or the costs associated with obtaining precise measurements on a large scale are prohibitive.

Multi-resolution learning algorithms serve as a solution for effectively integrating information from data collected over diverse temporal and spatial scales. Beyond addressing data availability constraints, multi-resolution methods can offer a reduction of computational costs during training, initially leveraging only lower-resolution data to capture general dynamics, and subsequently fine-tuning the models on high-resolution data.

To this end, a wide array of approaches has been proposed to tackle multi-resolution, multi-scale, or discretization-invariant learning. Among these, a natural framework that has gained prominence is neural operators, which are neural networks that learn mappings between function spaces, e.g., DeepONet, Fourier neural operator and others \citep{Lu_2021, fourier_2020, lno2023, wang2022discretization, RFB15a, seidman2022nomad}. %kontolati2023learning, borrel2023sound
Alternative methods for multi-resolution learning include encoder-decoder-based architectures \citep{ong2022iae, aboutalebi2022medusa}, graph and message-passing networks \citep{equer2023multi, liu2021multi}, and combinations of the above \citep{yang2022amgnet}. In a related vein, another body of research deals with conceptually similar multi-fidelity learning in DeepONets, which tackles the problem of combining data of varying quality, where high-quality samples are sparse \citep{lu2022multifidelity, howard2022multifidelity, de2023bi}.

%employing lower-quality data to capture general patterns and later tuning the network on high-quality samples to capture fine-grained details of the systems behavior.

In this work, we approach multi-resolution learning through the innate discretization-invariance property of DeepONets. We further propose a new architecture, DON-LSTM, which extends the %seminal 
architecture of the DeepONet with a long short-term memory network (LSTM), in order to capture temporal patterns in sequential outputs of the DeepONet. The main purpose of combining these two architectures is to leverage data of different resolutions in training, effectively increasing the feasible training set, as well as to assist the modeling of time-dependent evolution through explicit mechanisms of the LSTM. 

The remainder of this paper is structured as follows. First, we formulate the learning problem at hand. Next, we describe our proposed architecture and the training procedure. Finally, we present and discuss our experimental results on four non-linear partial differential equations (PDEs) and low- and high-resolution training sets of various sizes. Our main findings show that our proposed multi-resolution DON-LSTM achieves lower generalization error than its single-resolution counterparts, which require a much larger high-resolution sample size in order to achieve similar precision.

%Learning evolution of systes over time is important
% In real life data can be difficult to obtain
% This is why we want multiresolution method
% Different methods exist, cite some papers. 
% In this work we propose using the operator learning paradigm
% Some works on operators and multifidelity also exist
% We propose combining the DeepONet with LSTM and show that it works
% 

\section{Problem statement}

In this study, we learn the operator $\mathcal{N}$ that defines the evolution of a system over time starting from any given initial condition, i.e.:
\begin{equation}
    \mathcal{N}: u(x,t=0) \rightarrow G(u(y)),
\end{equation}
where $u(x,t=0)$ is the function defining the initial condition and $G(u(y))$ is the operator describing the evolution over time for any $y=(x,t)$, where $x$ and $t$ are the spatial and temporal coordinates of the system's trajectory.

In practice, $G(u(y))$ is observed at discretized fixed locations $\{(x_1,t_1),...,(x_m,t_n)\}$, resulting in a vector $[u(x_1, t_1), ..., u(x_m, t_n)] \in \mathbb{R}^{m\times n}$ where $m$ is the total number of spatial discretization points and $n$ is the total number of temporal discretization points that define the full trajectory. For the purpose of this study, we assume that for each system we have two datasets:
\begin{itemize}
    \item High-resolution set $D_H$ with $N_H$ samples of time resolution $\Delta t_{H}$,
    \item Low-resolution set $D_L$ of size $N_L$ samples of time resolution $\Delta t_{L}$.
\end{itemize}
For the four considered PDE-based examples in this work, we set $N_H=4\times N_L$ and $\Delta t_L = 5\times \Delta t_H$, while the spatial discretization is the same in both datasets. The multi-resolution network is trained on both datasets. The sizes of the training data are dependent on the complexity of the problem.

\section{Proposed Multi-resolution DON-LSTM (ours)}

\begin{figure}[h]
\begin{center}
\includegraphics[width=0.95\textwidth]{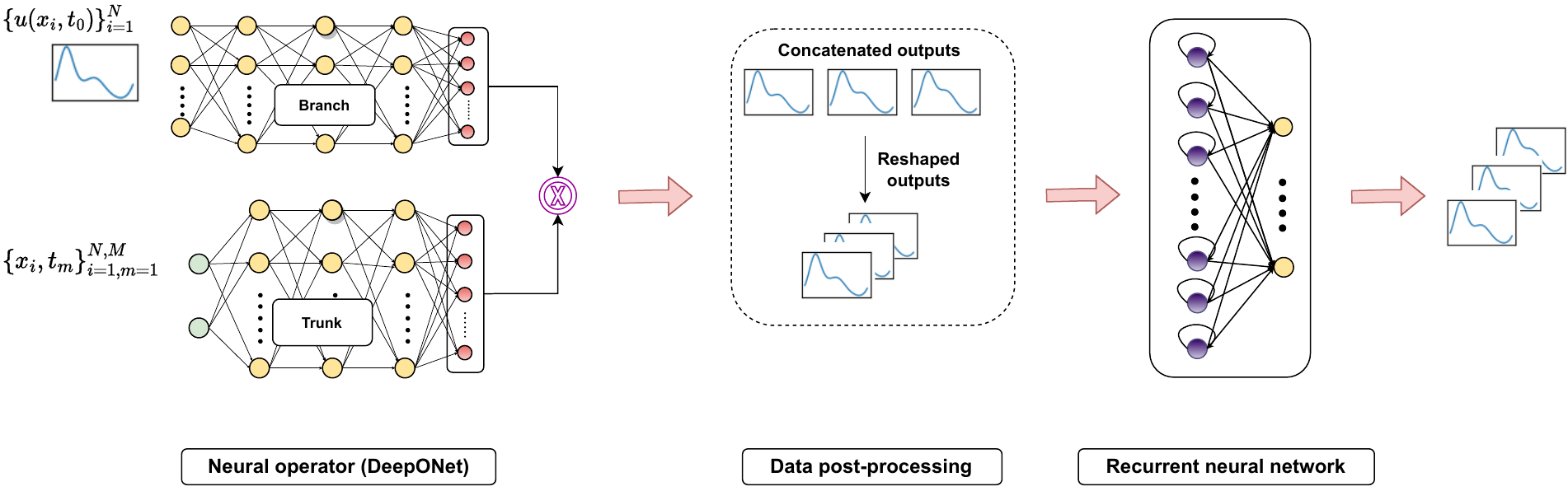}
\end{center}
\caption{The proposed architecture. In the first phase, the DeepONet maps the solution operator from the initial condition $u_{t=0}$ to solutions at later timesteps. During the DeepONet training, the output $u$ can be defined at arbitrary temporal locations. In the next two training stages, the outputs must have a fixed time interval $\Delta t$ to be compatible with the LSTM. The DeepONet output solutions are reshaped to be represented in a temporally sequential manner and processed by an LSTM. The LSTM lifts the dimension to the specified number of neurons and returns all hidden states. The last layer is a fully-connected dense layer bringing the embedding to the original size of the solution.}
\label{fig:proposed_architecture}
\end{figure}

The schematic representation of the proposed architecture is shown in \autoref{fig:proposed_architecture}. The architecture consists of a DeepONet followed by a reshaping layer, an LSTM layer, and a final feed-forward layer which brings the output back to the predefined spatial dimension of the solution. The architecture is set up such that the DeepONet outputs are fed as inputs to the LSTM. The DeepONet approximates the solution at locations that are determined by its inputs, which enables the initial discretization-invariant training. During the training of the LSTM, we impose the sequential nature of the outputs through these inputs and use the LSTM to process them as temporal sequences.

\subsection{Deep operator network (DeepONet)}
\label{deep_operator_explanations}

The DeepONet is an architecture based on the universal approximation theorem for operators \citep{chen1995universal} which combines the outputs of two deep neural networks (the "branch" and "trunk" networks). It is formulated as \citep{Lu_2021}:
\begin{equation}
    G(u)(y)\approx \mathop{\sum }\limits_{k=1}^{p}{\underbrace{{b}_{k}(u({x}_{1}),u({x}_{2}),\ldots ,u({x}_{m}))}_{\begin{array}{c}{\mathrm{branch}}\end{array}}}
\odot{\underbrace{{t}_{k}(y)}_{\begin{array}{c}{\mathrm{trunk}}\end{array}}},
\end{equation}
where ${b}_{k}$ and ${t}_{k}$ denote the output embeddings of the branch and the trunk network, respectively, and $\odot$ denotes the element-wise multiplication of the outputs of the two networks. The inputs to the branch network are the initial conditions, $u(t=0)$, discretized at $m$ spatial sensor locations $\{x_1,x_2, \ldots ,x_m\}$, and for the trunk network they are the locations, $y=(x,t)$, at which the operator $G(u)$ is evaluated. Additionally, the trunk network can incorporate periodic boundary conditions through simple feature expansion that consists of applying Fourier basis on the provided spatial locations, i.e., $x \rightarrow \cos(\frac{2\pi x}{P})$, and  $x \rightarrow \sin (\frac{2\pi x}{P})$, where $P$ is the period \citep{lu2022comprehensive}. 

%Neural operators learn nonlinear mappings between infinite-dimensional functional spaces on bounded domains, providing a unique simulation framework for the real-time prediction of multi-dimensional complex dynamics. Such models are discretization invariant, which means they share the same network parameters across different discretizations of the underlying functional data. In this study, we consider the performance the DeepONet proposed in \cite{Lu_2021}. 

%\noindent\textbf{DeepONet}: The deep operator network is based on the universal approximation theorem for operators \cite{chen1995universal} and employs two deep neural networks (branch and trunk network) to learn a family of PDEs and provide a discretization-invariant emulator, which allows for fast inference and low generalization error \cite{lanthaler2022error}. The branch network encodes the input function (the initial or boundary conditions, constant or variable coefficients, source terms) at fixed sensor points while the trunk network encodes the information related to the spatio-temporal coordinates of the output function. The output embeddings of the branch and the trunk networks are multiplied element-wise and summed over the neurons in the last layer of the networks (Einstein summation). 

%Describe the periodic boundary conditions

\subsection{Long short-term memory network (LSTM)}

LSTM is a type of recurrent neural network (RNN), which is a class of specialized neural networks for processing long sequences. In contrast to traditional feed-forward networks, RNNs employ a hidden state, which persists as the network iterates through data sequences and propagates the information from the previous states forward. LSTMs were developed as an extension to RNNs, aimed at mitigating the problem of vanishing gradients encountered in RNNs. Vanishing gradients occur in long sequences, as the hidden state is updated at every timestep, which leads to a large number of multiplications causing the gradients to approach zero. To address this, LSTM is equipped with a set of gates: the \textit{input gate} regulates the inflow of new information to the maintained hidden state, the \textit{forget gate} determines what proportion of the information should be discarded, and the \textit{output gate} controls the amount of information carried forward.

\subsection{Self-adaptive loss function}
In solving time-dependent PDEs, the main challenge lies in preserving the precision of the modeled solution over a long-time horizon.
%the solution at the later timesteps require stricter penalization over the early timesteps to preserve the wave shape. 
Introducing appropriately structured non-uniform penalization parameters can address this aspect. In all the experiments, we equip the networks with self-adaptive weights in the loss function, first introduced by \citet{mcclenny2020self}, which adapt during training and force the network to focus on the most challenging regions of the solution. We take the inspiration from \cite{kontolati2022influence}, where in turn self-adaptive weights considerably improve the accuracy prediction of discontinuities or non-smooth features in the solution. The self-adaptive weights are defined for every spatial and temporal location, and they are updated along with the network parameters during optimization. Specifically, the training loss is defined as:
\begin{equation}
    \mathcal L(\bm \theta, \bm \lambda) = \frac{1}{N}\sum_{i = 1}^{N} g(\bm\lambda)|u(x_i, t_i)-\hat{u}(x_i, t_i)|^2,
\end{equation}
where $(u(x_i, t_i)-\hat{u}(x_i, t_i))^2$ is the squared error between the reference $u$ and predicted value $\hat{u}$, and $g(\lambda)$ is a non-negative, strictly increasing self-adaptive mask function, and $\bm \lambda = \{\lambda_1, \lambda_2, \cdots \lambda_j\}$ are $j$ self-adaptive parameters, and $j$ is the total number of evaluation points.  Typically, in a neural network, we minimize the loss function with respect to the network parameters, $\boldsymbol{\theta}$. However, in this approach, we additionally maximize the loss function with respect to the trainable hyper-parameters using a gradient descent/ascent procedure. The modified objective function is defined as:
\begin{equation}
    \min_{\bm \theta} \max_{\bm \lambda}\mathcal L(\bm \theta, \bm \lambda).
\end{equation}
The self-adaptive weights are updated using the gradient descent method, such that
\begin{equation}
    \bm \lambda^{k+1} = \lambda^{k} + \eta_{\lambda}\nabla_{\bm \lambda}\mathcal L(\bm \theta, \bm \lambda),
\end{equation}
where $\eta_{\bm \lambda}$ is the learning rate of the self-adaptive weights and
\begin{equation}
    \nabla_{\lambda_i}\mathcal L = \left[g'(\lambda_i)(u_i (\xi)- \mathcal G_{\mathbf\theta}(\mathbf{v}_i)(\xi))^2\right]^{T}.
\end{equation}
Therefore, if $g(\lambda_i)>0$, $\nabla_{\lambda_i}\mathcal L$ would be zero only if the term $(u_i (\xi)- \mathcal G_{\mathbf\theta}(\mathbf{v}_i)(\xi))$ is zero. In this work, the self-adaptive weights are normalized to sum up to one after each epoch. The weights $\lambda$ are updated alongside the weights of the network during training.

\subsection{Training procedure}

%\begin{algorithm}
%\caption{The training phase of the baseline algorithm}\label{alg:baseline_train}
%\KwData{Observations $D=\{(t_1, \vec{x}^1, \vec{u}^1), \ldots, (t_N, \vec{x}^N, \vec{u}^N\}$}
%\KwData{Number of epochs $K$}
%\KwData{Batch size $M_b$}
%\KwData{Initial CNN $g_\theta$}
%\KwData{Loss function $\mathcal{L}_{g_\theta}$ defined in \eqref{eq:loss_ode}}
%\KwResult{Parameters $\theta$ for  $g_\theta$}
%\For{k in $1\ldots K$}{
%\For{batch in Batches}{
%    $B:=\{(u^{j_m}, u^{j_m+1}, t^{j_m})\}_{m=1}^{M_b} \gets \text{DrawRandomBatch}(D, M_b) $\;
%    Step using $\mathcal{L}_{g_\theta}(B)$ and $\nabla_\theta \mathcal{L}_{g_\theta}(B)$ 
%}
%}
%\end{algorithm}

%\noindent \textbf{Procedure:} Train:

Training is performed iteratively over a predefined number of epochs. The training is performed in minibatches, i.e., in each epoch the training data is divided into multiple smaller subsets. After calculating the loss on each batch, the gradients are calculated w.r.t.\ the loss function and the trainable network parameters (weights) are updated. The optimization is performed using the Adam optimizer and a predefined learning rate. The training of the multi-resolution DON-LSTM is performed in three stages:

\noindent \textbf{Step 1:} DeepONet training:
\begin{enumerate}
    \item Initialize the DeepONet with a set of weights $\bm \theta_{DON}$.
    \item Iteratively update $\bm \theta_{DON}$ on low-resolution data $D_L$ with a learning rate $lr_1$, saving weights $\bm \theta_{DON_i}$ at every $n_{freq}$ epochs.
    \item Compute the MSE loss on the predictions on validation data and choose the set of weights that correspond to the lowest loss.
\end{enumerate}

\noindent \textbf{Step 2:} LSTM training:
\begin{enumerate}
    \item Extend the DeepONet with the reshaping layer and the LSTM layer. 
    \item Initialize the LSTM with a set of weights $\bm \theta_{LSTM}$.
    \item Freeze $\bm \theta_{DON}$ (set as non-trainable parameters).
    \item Iteratively update $\bm \theta_{LSTM}$ on high-resolution data $D_H$, with a learning rate $lr_1$, saving models at every $n_{freq}$ epochs.
    \item Compute the MSE loss on the predictions on validation data and choose the set of weights that correspond to the lowest loss.
\end{enumerate}

\noindent \textbf{Step 3:} DON-LSTM training:

\begin{enumerate}
    \item Unfreeze $\bm \theta_{DON}$ (set as trainable parameters).
    \item Iteratively update $\bm \theta_{LSTM}$ on high-resolution data $D_H$ with a learning rate $lr_2$, where $lr_2 < lr_1$, saving models at every $n_{freq}$ epochs. 
    \item Compute the MSE loss on the predictions on validation data and choose the set of weights that correspond to the lowest loss.
\end{enumerate}

%\section{Partial differential equations data}
\section{Problems considered and data generation}
\label{sec:problems}
The performance of the proposed multi-resolution models is showcased on four infinite-dimensional non-linear dynamical systems. The numerical solutions to the Korteweg--de Vries, Benjamin--Bona--Mahony and Cahn--Hilliard equations were obtained through schemes that are second order in both space and time; the equations are spatially discretized using central finite differences and integrated in time using the implicit midpoint method. The solutions of the Burgers' equation were generated using PDEBench \citep{takamoto2022pdebench}. %In the following, the subscripts denote partial derivatives with respect to that variable. 
All PDEs are evaluated on the one-dimensional spatial domain $\Omega =[0,P]$, for different $P$, with periodic boundary conditions $u(P,t)=u(0,t)$ for all $t\geq 0$.
The training set for each PDE consists of data obtained from $N=N_H+N_L$ different initial conditions integrated over time $t=T$ with step size $\Delta t$. Example data are visualized in \autoref{fig:data_examples} and details of the time and space domains are given in \autoref{table:data_specifications}, both in \autoref{app:data}.

\subsection{Korteweg--de Vries equation}

The Korteweg--de Vries (KdV) equation \citep{Korteweg1895xli} is a non-linear dispersive PDE that describes the evolution of small-amplitude, long-wavelength systems in a variety of physical settings, such as shallow water waves, ion-acoustic waves in plasmas, and certain types of nonlinear optical waves. 
We consider a one-dimensional unforced case, which is given by:
\begin{equation}
\label{eq:kdv}
    \frac{\partial u}{\partial t} + \gamma \frac{\partial ^3 u}{\partial x^3} - \eta u \frac{\partial u}{\partial x} = 0,
\end{equation}
%\begin{equation}
%\label{eq:kdv}
%    u_t  - \eta uu_x + \gamma u_{xxx} = 0,
%\end{equation}
where $u:=u(x,t)$ is the height of the wave at position $x$ at time $t$, and $\eta=6$ and $\gamma=1$ are chosen real-valued scalar parameters.% $x$ is the spatial and $t$ is the time dimension. %The subscripts denote partial derivatives with respect to that variable.

The initial conditions are each a sum of two solitons (solitary waves), \textit{i.e.}:
\begin{equation}
u(x,0) =  \sum_{i=1}^2 { 2 k_i^2  \text{sech}^2 \left( k_i ( (x+\frac{P}{2}-Pd_i) \% P - \frac{P}{2} )\right)},
\end{equation}
where $\mathrm{sech}$ stands for hyperbolic secant ($\frac{1}{\mathrm{cosh}}$), $P$ is the period in space, $\%$ is the modulo operator, $i=\{1,2\}$, $k_i\in (0.5, 1.0)$ and $d_i\in (0, 1)$ are coefficients that determine the height and location of the peak of a soliton, respectively. For each initial state $t=0$ in the training data, these coefficients are drawn randomly from their distribution.

\subsection{Benjamin--Bona--Mahony equation}

The Benjamin--Bona--Mahony (BBM) equation was derived as a higher-order improvement on the KdV equation and includes both non-linear and dispersive effects \citep{Peregrine1966calculations, Benjamin1972model}. It is used for studying a broader range of wave behaviors, including wave breaking and dispersion, and is given by:
\begin{equation}
\frac{\partial u}{\partial t} - \frac{\partial^3 u}{\partial x^2 \partial t} + \frac{\partial}{\partial x} \left( u  + \frac{1}{2}u^2 \right) = 0.
\label{eq:bbm} 
\end{equation}
%\begin{equation}
%u_t - u_{xxt} = - \frac{\partial}{\partial x} \left( u  + \frac{1}{2}u^2 \right),
%\label{eq:bbm} 
%\end{equation}
The chosen initial conditions are superpositions of two soliton waves of the following shape:
\begin{equation}
u_i(x,0) = \sum_{i=1}^2 {3 (c_i-1) \text{sech}^2\left(\frac{1}{2} \sqrt{1-\frac{1}{c_i}} (x+\frac{P}{2}-P d_i) \% P - \frac{P}{2}\right)},
\end{equation}
where $c_i \in (1,3)$ and $d_i\in(0,1)$ are coefficients that determine the height and location of
the peak of a soliton, respectively.

\subsection{Cahn--Hilliard equation}

The Cahn--Hilliard equation is used to describe phase separation with applications to materials science and physics \citep{Cahn1958free}. It is expressed as:
\begin{equation}
\frac{\partial u}{\partial t}  - \frac{\partial^2}{\partial x^2}(\nu u + \alpha u^3 + \mu \frac{\partial^2 u}{\partial x^2}) = 0,
\label{eq:cahn_hilliard}
\end{equation}
%\begin{equation}
%u_t  - (\nu u + \alpha u^3 + \mu u_{xx})_{xx} = 0,
%\label{eq:cahn_hilliard}
%\end{equation}
where we set $\nu = -0.01$, $\alpha = 0.01$ and $\mu = -0.00001$.

The initial conditions are superpositions of sine and cosine waves, i.e. $u(x,0)=u_1+u_2$ with:
%    u0 = a0 * np.sin(k0*np.pi / x_max*grid_x)
%    u0+= a1 * np.cos(k1*np.pi / x_max*grid_x) 
%    u0+= a2 * np.sin(k2*np.pi / x_max*grid_x) 
%    u0+= a3 * np.cos(k3*np.pi / x_max*grid_x)
\begin{equation}
u_i(x,0) = a_i \text{sin}(k_i \frac{2\pi}{P}x) + b_i \text{cos}(j_i \frac{2\pi}{P}x),
\end{equation}
where $a_i, b_i \in (0,0.2)$ and $k_i, j_i$ are integers and $j_i, k_i \in [1,6]$.

\subsection{Viscous Burgers' equation}

The viscous Burgers' equation \citep{Bateman1915some, Burgers1948mathematical} describes the behavior of waves and shocks in a viscous fluid or gas. It is given by:
\begin{equation}
    \frac{\partial u }{\partial t} + \frac{\partial}{\partial x}\left(\frac{u^2}{2}\right)= \frac{\nu}{\pi}\frac{\partial u^2}{\partial ^2 x},
\end{equation}
%\begin{equation}
%    u_t+ \left(\frac{u^2}{2}\right)_x= \frac{\nu}{\pi} u_{xx},
%\end{equation}
where $x \in (0,1)$, $t\in (0,2]$, and $\nu=0.001$ is the diffusion coefficient.

The initial conditions are given by the superposition of sinusoidal waves:
%\begin{equation}
%    u(x,0)=u_0(x), 
%    x \in (0,1),
%\end{equation}
\begin{equation}
\label{init_sin_waves}
    u(x, 0) = \sum_{k_i=k_1,...,k_N} A_i \sin(k_ix+\phi_i),
\end{equation}
where $k_i=2\pi{n_i}/P$ are coefficients where ${n_i}$ are arbitrarily selected integers in $[1,n_{max}]$. $N$ is the integer determining how many waves are added, $A_i$
is a random float number uniformly chosen in $(0, 1)$, and $\phi_i$
is the randomly chosen phase
in $(0, 2\pi)$. 

\section{Experimental results}

We compare the performance of our multi-resolution DON-LSTM against five benchmark networks discussed in \autoref{sec:problems} using the relative squared error (RSE), the mean average error (MAE) and the root mean squared error (RMSE), described in \autoref{app:eval_metrics_explained}. The average values of these error metrics are presented in \autoref{tab:errors_allproblems}, and the log values of RSE against the increasing number of high-resolution training samples are shown in \autoref{fig:rse_resolution}. Due to space considerations, detailed tables with the evaluation of the models trained on distinct sample sizes are available in \autoref{app:performance_metrics}.
 All evaluations are performed on high-resolution test samples of size $N_{test}=1000$.

\subsection{Benchmark models}

%\begin{table}[h]
%\caption{Benchmark models}
%\label{table:benchmark_models}
%\begin{center}
%\begin{tabular}{ll}
%\multicolumn{1}{c}{\bf Model}  &\multicolumn{1}{c}{\bf Purpose of comparison}
%\\ \hline \\
%DON (low-res.)         & Test the impact of large low-resolution training data on prediction. \\
%DON (high-res.)             & Test the impact of high-resolution data on prediction. \\
%LSTM (high-res.)             & Test the impact of including explicit mechanisms to capture temporal dependencies in the output. \\
%DON-LSTM (high-res.) & Test the performance of the proposed architecture without training on low-resolution data. \\
%\end{tabular}
%\end{center}
%\end{table}

The benchmark models used to evaluate the performance of the multi-resolution DON-LSTM (DON-LSTM ($D_H$, $D_L$)) are the following:
\begin{itemize}
    \item \textbf{DON ($D_L$)}: The vanilla DeepONet trained on $N_L$ low-resolution data,
    \item \textbf{DON ($D_H$)}: The vanilla DeepONet trained on $N_H$ high-resolution data,
    \item \textbf{DON ($D_H, D_L$)}: The vanilla DeepONet trained on $N_L+N_H$ multi-resolution data (both datasets),
    \item \textbf{LSTM ($D_H$)}: An architecture trained on $N_H$ high-resolution data, consisting of one dense layer lifting the input to a dimension [-1, $mn$], a reshaping layer (into [-1, $m$, $n$]) followed by an LSTM layer, where $m$ is the spatial and $t$ is the temporal dimension,
    \item \textbf{DON-LSTM ($D_H$)}: The proposed architecture trained only on $N_H$ high-resolution data.
\end{itemize}

The models vary by the amount and granularity of the training data. We take advantage of the discretization-invariance of deep neural operators and train the vanilla DeepONet and DON-LSTM on multi-resolution data (in the case of DON-LSTM only the DeepONet layers are trained with multi-resolution). In this problem formulation, the LSTM  implicitly learns $\Delta t$ and therefore has to be trained at the resolution used in testing (here, high-resolution). In the following, we refer to specific models by the data resolution used in their training and the models' name, e.g., \textit{high-resolution DON}.

\subsection{Generalization performance}
\label{results}

\begin{figure}[ht]
%\vskip 0.2in
\begin{center}

\includegraphics[width=\textwidth]{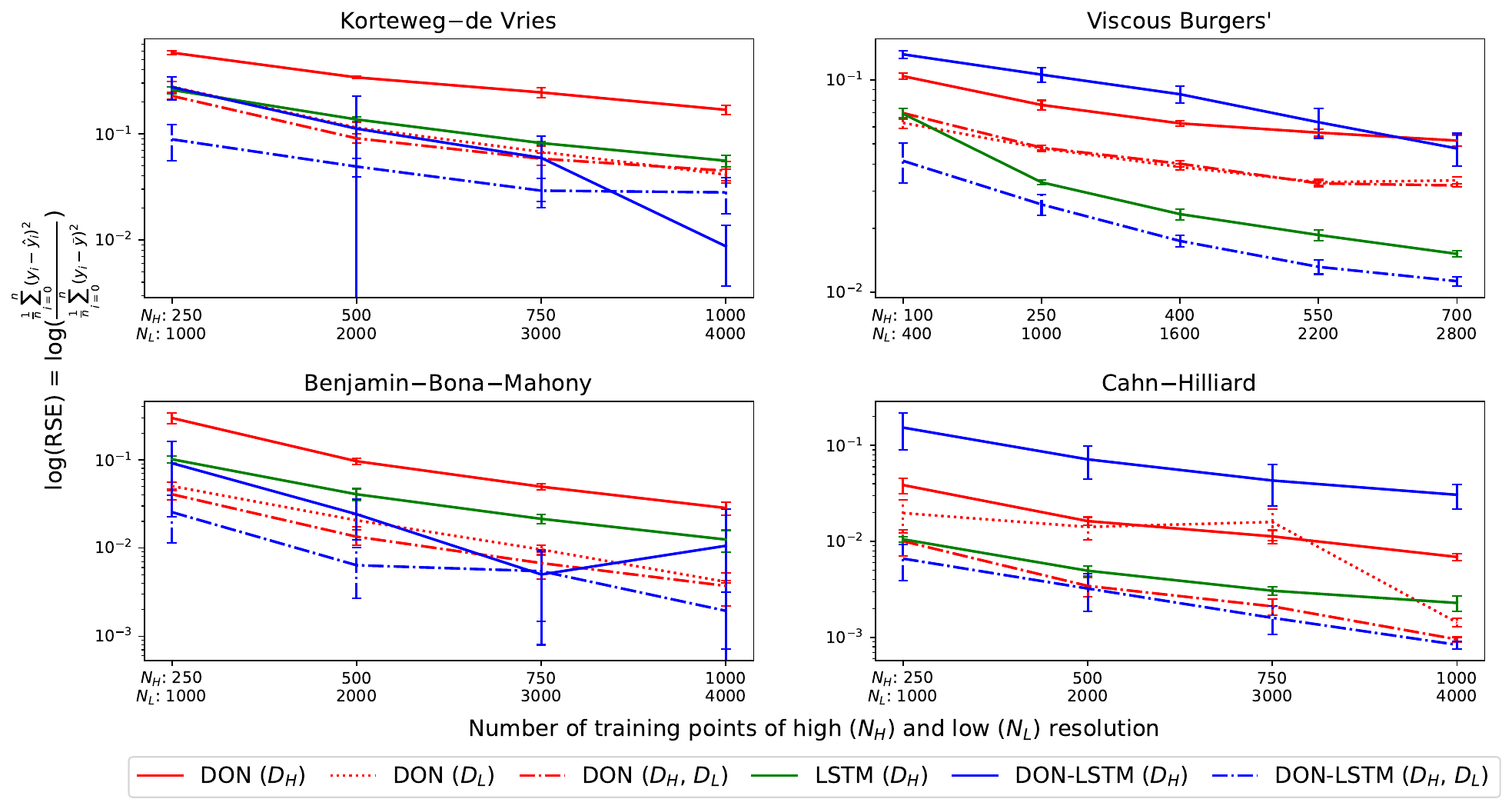}

%\centerline{
%\begin{subfigure}{0.5\textwidth}
%\includegraphics[width=\textwidth]{figures/rse_comparison_kdv_ylog_6models2.pdf}
%\caption{Korteweg--de Vries equation.}
%\label{subfig:rse_resolution_kdv}
%\end{subfigure}
%\hfill
%\begin{subfigure}{0.5\textwidth}
%\includegraphics[width=\textwidth]%{figures/rse_comparison_burgers_legend_ylog_6models2.pdf}
%\caption{Burgers' equation.}
%\label{subfig:rse_resolution_burgers}
%\end{subfigure}
%}
%\vskip 0.2in
%\centerline{
%\begin{subfigure}{0.5\textwidth}
%\includegraphics[width=\textwidth]{figures/rse_comparison_bbm_legend_ylog_font.pdf}
%\caption{Benjamin--Bona--Mahony equation.}
%\label{subfig:rse_resolution_bbm}
%\end{subfigure}
%\hfill
%\begin{subfigure}{0.5\textwidth}
%\includegraphics[width=\textwidth]%{figures/rse_comparison_cahn_hilliard_legend_ylog_6models2.pdf}
%\caption{Cahn--Hilliard equation.}
%\label{subfig:rse_resolution_cahn_hilliard}
%\end{subfigure}
%}
\caption{Model performance vs. amount of training samples. The {$y$-axis} shows the relative squared error (log values) for predictions on high-resolution test data, while the $x$-axis indicates the number of high- and low-resolution training samples. The figure compares the performance of three models: DON (red) and DON-LSTM (blue) and LSTM (green), with variations of training data denoted by $D_{\square}$ and the line pattern ($D_H$ and solid line for high-resolution, $D_L$ and dotted line for low-resolution, and $D_H$, $D_L$ and dash-dotted line for multi-resolution).  The error bars indicate the $95\%$ confidence interval calculated over the five trained models (two standard deviations of the error).}

%vs. the number of high-resolution training samples. 

%The vanilla DeepONet (red) and DON-LSTM (blue) can be trained on multi-resolution data, while the vanilla LSTM (green) requires single-resolution data. The models are trained on high-resolution data (solid line), low-resolution data (dotted line) or multi-resolution data (dash-dotted line, both high- and low-resolution), depending on the model's data requirements.}
\label{fig:rse_resolution}
\end{center}
\vskip -0.2in
\end{figure}

We evaluate the performance of our models grouping them by the number of samples used in their training (\autoref{fig:rse_resolution}). As expected, increasing the sample size leads to a reduction in the generalization error for all models. In nearly all cases, we also see that the multi-resolution DON-LSTM achieves the lowest error, followed by the multi-resolution DON for the KdV, BBM and Cahn--Hilliard equations, and the high-resolution LSTM for the Burgers' equation.

% We see a discrepancy between the performance of high-resolution DON and DON-LSTM - while they both dont manage to achieve low error, the DON-LSTM is better in KdV and BBM, and the DON is better in Burgers' and Cahn-Hilliard. 

Our main findings can be summarized into the following:

\begin{enumerate}
    \item The multi-resolution DON-LSTM generally achieves the lowest generalization error out of the five benchmarks.
    \item In order to achieve similar accuracy with single-resolution methods (such as the vanilla LSTM) we need significantly more high-resolution training samples than for multi-resolution DON-LSTM.
    \item In multiple cases the DON trained on larger amount of lower-resolution data obtains better results than the DON trained on fewer samples of high-resolution data.
    \item While the DeepONet itself achieves reasonable performance, the time-dependent architecture is crucial for capturing long-time dynamics, as is evident by the superior performance of DON-LSTM and the vanilla LSTM. 

\end{enumerate}

\begin{table}[ht]
\centering
\begin{tabular}{lllccc}
\hline 
\textbf{Model} & \multicolumn{2}{l}{\textbf{Resolution}}  &              \textbf{MAE} &             \textbf{RMSE} &              \textbf{RSE} \\
\midrule 
\multicolumn{6}{c}{\textbf{Korteweg--de Vries equation}} \\
\hline 

     \textbf{DON} &    $\Delta t= 0.025$ &   (high) &          0.190$\pm$0.077 &          0.321$\pm$0.086 &          0.333$\pm$0.179 \\
     \textbf{DON} &    $\Delta t= 0.125$ &      (low) &          0.094$\pm$0.059 &          0.189$\pm$0.080 &          0.125$\pm$0.106 \\
     \textbf{DON} &   $\Delta t= \{0.025, 0.125\}$  &    (multi) &          0.083$\pm$0.045 &          0.175$\pm$0.068 &          0.105$\pm$0.083 \\
\textbf{DON-LSTM} &    $\Delta t= 0.025$ &   (high) &          0.086$\pm$0.052 &          0.168$\pm$0.102 &          0.113$\pm$0.115 \\
\textbf{DON-LSTM} &   $\Delta t= \{0.025, 0.125\}$  &    (multi) & \textbf{0.042$\pm$0.018} & \textbf{0.122$\pm$0.034} & \textbf{0.049$\pm$0.028} \\
    \textbf{LSTM} &    $\Delta t= 0.025$ &   (high) &          0.067$\pm$0.032 &          0.200$\pm$0.068 &          0.133$\pm$0.090 \\

\midrule 

\multicolumn{6}{c}{\textbf{Viscous Burgers' equation}} \\
\hline 

     \textbf{DON} &    $\Delta t= 0.01$ &    (high) &          0.114$\pm$0.016 &          0.168$\pm$0.024 &          0.070$\pm$0.021 \\
     \textbf{DON} &     $\Delta t= 0.05$ &    (low) &          0.089$\pm$0.013 &          0.132$\pm$0.018 &          0.043$\pm$0.012 \\
     \textbf{DON} &    $\Delta t= \{0.01, 0.05\}$ &   (multi) &          0.087$\pm$0.014 &          0.129$\pm$0.019 &          0.044$\pm$0.016 \\
\textbf{DON-LSTM} &     $\Delta t= 0.01$ &   (high) &          0.111$\pm$0.034 &          0.186$\pm$0.037 &          0.087$\pm$0.033 \\
\textbf{DON-LSTM} &    $\Delta t= \{0.01, 0.05\}$ &   (multi) & \textbf{0.049$\pm$0.017} & \textbf{0.092$\pm$0.025} & \textbf{0.022$\pm$0.012} \\
    \textbf{LSTM} &    $\Delta t= 0.01$ &    (high) &          0.059$\pm$0.028 &          0.110$\pm$0.036 &          0.032$\pm$0.022 \\
\midrule 

\multicolumn{6}{c}{\textbf{Benjamin--Bona--Mahony equation}} \\
\hline 

     \textbf{DON} &    $\Delta t = 0.075$ &   (high) &          0.278$\pm$0.191 &          0.533$\pm$0.284 &          0.118$\pm$0.123 \\
     \textbf{DON} &     $\Delta t = 0.375$ &   (low) &          0.091$\pm$0.055 &          0.227$\pm$0.118 &          0.021$\pm$0.021 \\
     \textbf{DON} &     $\Delta t = \{0.075, 0.375\}$ &    (multi) &          0.077$\pm$0.055 &          0.191$\pm$0.100 &          0.016$\pm$0.017 \\
\textbf{DON-LSTM} &    $\Delta t = 0.075$ &   (high) &          0.104$\pm$0.068 &          0.264$\pm$0.175 &          0.033$\pm$0.040 \\
\textbf{DON-LSTM} &      $\Delta t = \{0.075, 0.375\}$ &   (multi) & \textbf{0.045$\pm$0.030} & \textbf{0.151$\pm$0.084} & \textbf{0.010$\pm$0.011} \\
    \textbf{LSTM} &    $\Delta t = 0.075$ &   (high) &          0.107$\pm$0.061 &          0.332$\pm$0.155 &          0.044$\pm$0.040 \\

\midrule 

\multicolumn{6}{c}{\textbf{Cahn--Hilliard equation}} \\
\hline 

     \textbf{DON} &   $\Delta t = 0.02$  &     (high) &          0.041$\pm$0.018 &          0.068$\pm$0.026 &          0.018$\pm$0.014 \\
     \textbf{DON} &   $\Delta t = 0.1$  &      (low) &          0.038$\pm$0.019 &          0.056$\pm$0.024 &          0.013$\pm$0.008 \\
     \textbf{DON} &   $\Delta t = \{0.02, 0.1\}$ &  (multi) &          0.018$\pm$0.010 &          0.031$\pm$0.016 &          0.004$\pm$0.004 \\
\textbf{DON-LSTM} &     $\Delta t = 0.02$  &   (high) &          0.076$\pm$0.036 &          0.137$\pm$0.050 &          0.075$\pm$0.055 \\
\textbf{DON-LSTM} &   $\Delta t = \{0.02, 0.1\}$ &   (multi) & \textbf{0.014$\pm$0.007} & \textbf{0.027$\pm$0.012} & \textbf{0.003$\pm$0.003} \\
    \textbf{LSTM} &    $\Delta t = 0.02$  &    (high) &          0.016$\pm$0.006 &          0.036$\pm$0.013 &          0.005$\pm$0.004 \\

\hline 

\end{tabular}
\caption{The mean and the standard deviation of the prediction errors of all models. Each reported value aggregates the mean error values across all models trained on different sample sizes, where each model has been trained five times (i.e., the mean and standard deviation of the values reported in \autoref{app:performance_metrics}).}
\label{tab:errors_allproblems}
\end{table}

\section{Discussion}

In all our experiments, the multi-resolution DON-LSTM achieved the lowest generalization error, while requiring fewer high-resolution training samples than its benchmarks. This advantage stems from the combination of two factors: the utilization of a larger training dataset, which encompasses both high- and low-resolution samples, and the integration of LSTM mechanisms that facilitate capturing the temporal evolution of the systems. We specifically observe that the inclusion of LSTM mechanisms is beneficial, as evidenced by the multi-resolution DON-LSTM outperforming the vanilla DON trained on multi-resolution data. Additionally, the inclusion of low-resolution data in early training contributes to the improvement of the prediction, as seen in the superior performance of multi-resolution DON-LSTM as compared to the vanilla LSTM and single-resolution DON-LSTM, as well as the superior performance of the multi-resolution DON in comparison to both single-resolution DONs. 

When low-resolution data was not used in training, the comparison between the architectures was inconclusive, i.e., the high-resolution DON-LSTM outperformed its vanilla counterparts only in two experiments. We attribute it to the fact that DON-LSTM is comprised of a larger number of parameters, and a small training sample is not sufficient to effectively train the network, leading to under/over-fitting. We can also see that the vanilla DON struggles with adjusting all its parameters on a small sample, which becomes apparent through the fact that the model trained on a low-resolution data achieves better performance than when trained on the high-resolution data (regardless of being tested on high-resolution). This means that the inclusion of low-resolution data in early training is essential for the good performance of the proposed architecture.

\section{Limitations and Future work}

While DeepONets possess the discretization-invariance property in the output function, %which is heavily desired when working with multi-resolution time-series data, 
they require the input data to be defined at fixed locations. This problem is addressed in the literature through the employment of an encoder-decoder architecture integrated with the DeepONet \citep{ong2022iae, zhang2022multiauto}. %oommen2022learning, goswami2023learning
We also note that the framework that is limited to discretization-invariant output is sufficient for applications where the system behaviour is modeled from a single-resolution input, e.g., an initial condition. For cases when multi-resolution input is desired, we highlight the existence of other neural operator architectures such as the Fourier neural operator \citep{fourier_2020}, or the Laplace neural operator \citep{lno2023}.  

In addition, we note that LSTM is specifically suited and limited to sequential data. We see this as an opportunity for future studies, in which the DeepONet can be extended with architectures appropriate for different types of data, for example convolutional neural networks in case of image data, or transformers for data governed by less-regular temporal patterns.

\section{Conclusions}

Our proposed architecture, DON-LSTM, seamlessly integrates the strengths of its two composites: the discretization invariance of the DeepONet and the improved sequential learning with the memory-preserving mechanisms of the LSTM. 
We have demonstrated that these properties can be leveraged to incorporate multi-resolution data in training, as well as to capture the intricate dynamics of time-dependent systems, leading to significantly improved predictive accuracy in modeling of the systems' evolution over long-time horizons.
%The new architecture enables the incorporation of multi-resolution data during training and empowers the model to maintain high predictive accuracy over long-time horizons through the mechanisms that capture the intricate dynamics of sequential data. 
Our experiments clearly demonstrate the efficacy of our approach in creating accurate, high-resolution models even with limited training data available at fine-grained resolution. Moreover, the synergistic effect of our proposed architecture makes it an apt choice for real-world scenarios, promising substantial enhancements in prediction quality. This work not only advances the understanding and utilization of multi-resolution data in sequential analysis but also provides valuable insights for future research and applications.

%We have introduced a new architecture: DON-LSTM, which allows to include data at multiple resolutions in training, while achieving high predictive accuracy on sequential data. Through our experiments, we have shown that our proposed architecture can help achieve accurate high-resolution predictions in situations where high-resolution training data is limited. 

%combines the DeepONet and LSTM architectures. 
%Through this, we were able to effectively integrate multi-resolution training via the DeepONet and introduce a helpful mechanisms to capture long-term dynamics of the learned systems through the LSTM.
%We have shown that DeepONet can be used to exploit different resolution data during training, while by the proposed integration, we can still use architectures that are natural choices for real-world scenarios data and improve the predictions.
%It doesnt have to be LSTM, it can be another natural architecture

\section{Acknowledgement}

For KM and SRS, this work is based upon the support from the Research Council of Norway under project SFI NorwAI 309834. Furthermore, KM is also supported by the PhD project grant 323338 Stipendiatstilling 17 SINTEF (2021-2023). For SG and GEK, this work was supported by the U.S. Department of Energy, Advanced Scientific Computing Research program, under the Scalable, Efficient and Accelerated Causal Reasoning Operators, Graphs and Spikes for Earth and Embedded Systems (SEA-CROGS) project, DE- SC0023191. Furthermore, the authors would like to acknowledge the computing support provided by the computational resources and services at the Center for Computation and Visualization (CCV), Brown University where all the experiments were carried out, as well as the invaluable contribution in generating the training data by Sølve Eidnes.

\section{Reproducibility}

The code for reproducing the results is available in  \href{https://github.com/katarzynamichalowska/DON_LSTM}{GitHub repository} \citep{michalowska2023donlstm}. The code includes the default parameters to generate the models and the data processing pipeline used in this paper. The details of the used architectures, training setup and data processing steps are also specified in \autoref{app:architectures} and \autoref{appendix:normalization}.

\bibliography{iclr2024_conference}
\bibliographystyle{iclr2024_conference}

\appendix
%You may include other additional sections here.
\FloatBarrier

\newpage
\section{Partial differential equations data}

\label{app:data}

\begin{figure}[ht]
%\vskip 0.2in
\begin{center}
\centerline{
\begin{subfigure}{0.5\columnwidth}
\includegraphics[width=\columnwidth]{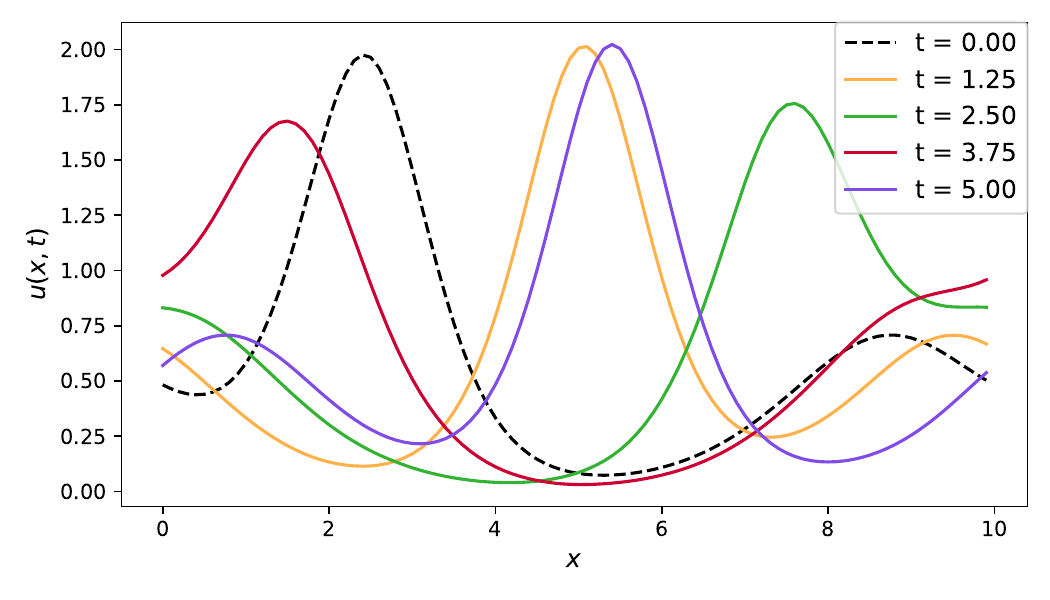}
\caption{Korteweg--de Vries equation.}
\label{subfig:data_kdv}
\end{subfigure}
\hfill
\begin{subfigure}{0.5\columnwidth}
\includegraphics[width=\columnwidth]{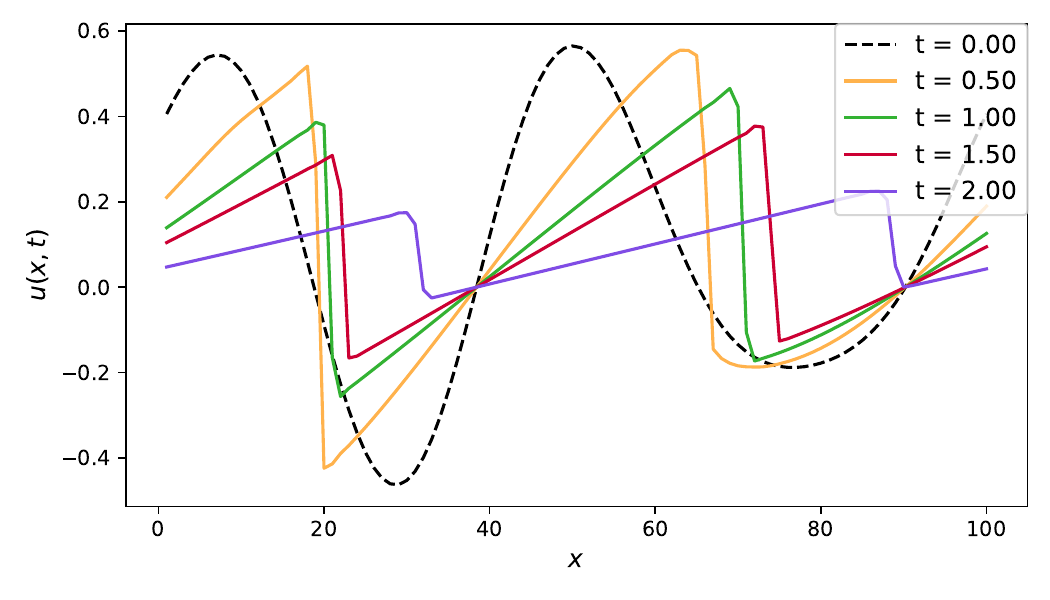}
\caption{Viscous Burgers' equation.}
\label{subfig:data_burgers}
\end{subfigure}
}
\vskip 0.2in
\centerline{
\begin{subfigure}{0.5\columnwidth}
\includegraphics[width=\columnwidth]{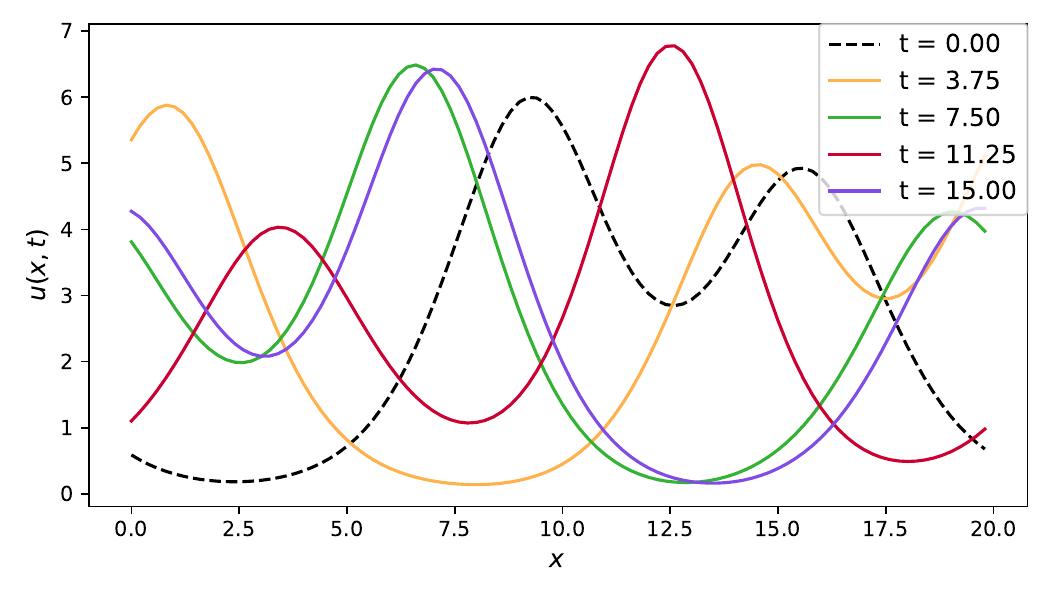}
\caption{Benjamin--Bona--Mahony equation.}
\label{subfig:data_bbm}
\end{subfigure}
\hfill
\begin{subfigure}{0.5\columnwidth}
\includegraphics[width=\columnwidth]{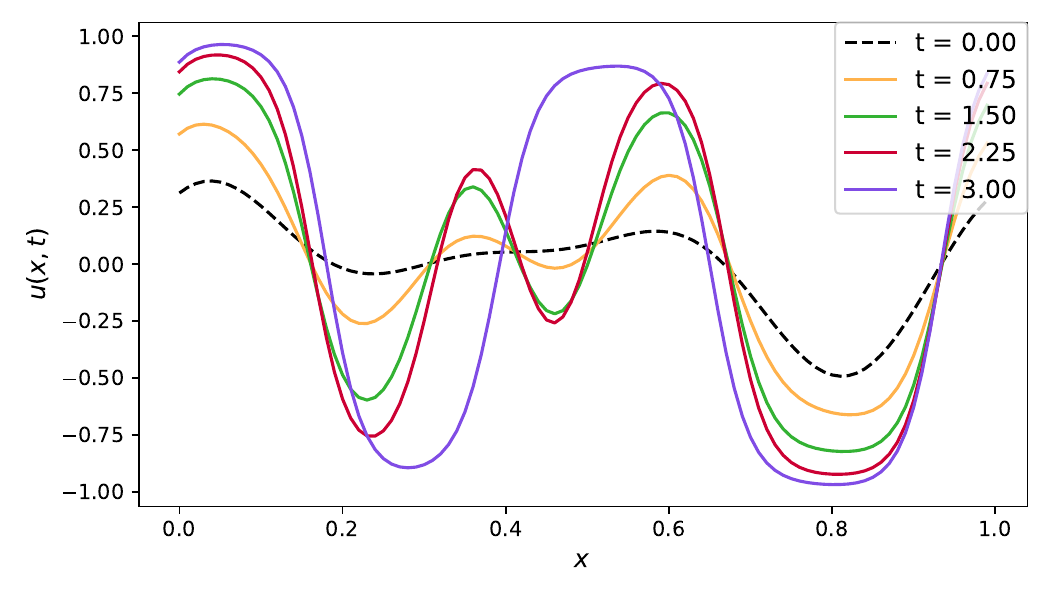}
\caption{Cahn--Hilliard equation.}
\label{subfig:data_cahn_hilliard}
\end{subfigure}
}
\vskip 0.2in
\caption{Random samples from training data. Each dataset consists of multiple trajectories starting from different initial conditions.}
\label{fig:data_examples}
\end{center}
\vskip -0.2in
\end{figure}

\begin{table}[h]
  \centering
\begin{tabular}{l|l|l|l|l}
\hline
\textbf{Equation}  &     $N$ &       \textbf{$t_{H}$} &             \textbf{$t_{L}$} & \textbf{$x$}\\
%\midrule
\hline
\multirow{3}{*}{Korteweg--de Vries} &  &\tabitem $t \in [0, 5]$  &  \tabitem $t \in [0, 5]$ &  \tabitem $\Omega \in [0, 10]$  \\

 &10000 & \tabitem $\Delta t = 0.025$ &  \tabitem $\Delta t = 0.125$ &  \tabitem $\Delta x = 0.1$ \\
 && \tabitem 201 points &  \tabitem 41 points &  \tabitem 100 points \\\hline

\multirow{3}{*}{Viscous Burgers'} & & \tabitem $t\in[0,2]$  & \tabitem $t\in[0,2]$ &  \tabitem $\Omega \in [-1, 1]$  \\
 &5000 &  \tabitem $\Delta t = 0.01$ &  \tabitem $\Delta t = 0.05$ & \tabitem $\Delta x = 0.02$\\
 & & \tabitem 201 points & \tabitem 41 points &  \tabitem 100 points  \\\hline

\multirow{3}{*}{Benjamin--Bona--Mahony} & & \tabitem $t \in [0,15]$& \tabitem $t \in [0,15]$& \tabitem $\Omega \in [0,20]$\\
 & 10000 & \tabitem $\Delta t = 0.075$& \tabitem $\Delta t =0.375$ & \tabitem$\Delta x=0.2$\\
 & & \tabitem 201 points & \tabitem 41 points &\tabitem 100  points\\\hline

 \multirow{3}{*}{Cahn--Hilliard} & &  \tabitem $t \in [0,3]$& \tabitem $t \in [0,3]$& \tabitem $\Omega \in [0,1]$\\
 & 10000 & \tabitem $\Delta t = 0.02$& \tabitem $\Delta t = 0.1$ & \tabitem$\Delta x=0.01$\\
 & & \tabitem  151 points & \tabitem  31 points &\tabitem   100 points\\
 \hline
\end{tabular}
\caption{Data available: data size, i.e. number of samples ($N$), time domain of high-resolution ($t_{H}$) and low-resolution data ($t_{L}$), spatial domain ($x$). We only use a small subset of this data in training.}
\label{table:data_specifications}
\end{table}

\FloatBarrier

\newpage
\section{Generalization performance}

\label{app:performance_metrics}

\begin{table}[ht]
  \centering
\begin{tabular}{lllccc}
\multicolumn{6}{c}{\textbf{Korteweg--de Vries equation}} \\
\hline 
\textbf{Model} & \multicolumn{2}{l}{\textbf{Resolution}}  &              \textbf{MAE} &             \textbf{RMSE} &              \textbf{RSE} \\
\hline 
\multicolumn{6}{l}{$N_H = 250$, $N_L = 1000$} \\
\hline 

     \textbf{DON} &            $\Delta t = 0.025$ &     (high) &          0.298$\pm$0.003 &          0.434$\pm$0.005 &          0.580$\pm$0.014 \\
     \textbf{DON} &            $\Delta t = 0.125$ &      (low) &          0.179$\pm$0.008 &          0.300$\pm$0.008 &          0.277$\pm$0.016 \\
     \textbf{DON} & $\Delta t = \{0.025, 0.125\}$ &    (multi) &          0.149$\pm$0.012 &          0.273$\pm$0.005 &          0.227$\pm$0.008 \\
\textbf{DON-LSTM} &            $\Delta t = 0.025$ &     (high) &          0.152$\pm$0.011 &          0.298$\pm$0.018 &          0.274$\pm$0.034 \\
\textbf{DON-LSTM} & $\Delta t = \{0.025, 0.125\}$ &    (multi) & \textbf{0.065$\pm$0.005} & \textbf{0.169$\pm$0.016} & \textbf{0.089$\pm$0.016} \\
    \textbf{LSTM} &            $\Delta t = 0.025$ &     (high) &          0.112$\pm$0.006 &          0.290$\pm$0.005 &          0.259$\pm$0.010 \\

\hline 
\multicolumn{6}{l}{$N_H = 500$, $N_L = 2000$} \\
\hline 
     \textbf{DON} &            $\Delta t = 0.025$ &     (high) &          0.192$\pm$0.011 &          0.332$\pm$0.002 &          0.339$\pm$0.005 \\
     \textbf{DON} &            $\Delta t = 0.125$ &      (low) &          0.084$\pm$0.002 &          0.193$\pm$0.006 &          0.114$\pm$0.007 \\
     \textbf{DON} & $\Delta t = \{0.025, 0.125\}$ &    (multi) &          0.075$\pm$0.004 &          0.169$\pm$0.004 &          0.091$\pm$0.005 \\
\textbf{DON-LSTM} &            $\Delta t = 0.025$ &     (high) &          0.089$\pm$0.025 &          0.185$\pm$0.050 &          0.112$\pm$0.058 \\
\textbf{DON-LSTM} & $\Delta t = \{0.025, 0.125\}$ &    (multi) & \textbf{0.044$\pm$0.002} & \textbf{0.126$\pm$0.006} & \textbf{0.049$\pm$0.005} \\
    \textbf{LSTM} &            $\Delta t = 0.025$ &     (high) &          0.069$\pm$0.003 &          0.211$\pm$0.003 &          0.136$\pm$0.003 \\

\hline

\multicolumn{6}{l}{$N_H = 750$, $N_L = 3000$} \\
\hline 

     \textbf{DON} &            $\Delta t = 0.025$ &     (high) &          0.149$\pm$0.004 &          0.282$\pm$0.008 &          0.245$\pm$0.013 \\
     \textbf{DON} &            $\Delta t = 0.125$ &      (low) &          0.066$\pm$0.005 &          0.148$\pm$0.005 &          0.068$\pm$0.004 \\
     \textbf{DON} & $\Delta t = \{0.025, 0.125\}$ &    (multi) &          0.059$\pm$0.003 &          0.137$\pm$0.005 &          0.058$\pm$0.004 \\
\textbf{DON-LSTM} &            $\Delta t = 0.025$ &     (high) &          0.077$\pm$0.010 &          0.138$\pm$0.021 &          0.059$\pm$0.018 \\
\textbf{DON-LSTM} & $\Delta t = \{0.025, 0.125\}$ &    (multi) & \textbf{0.034$\pm$0.002} & \textbf{0.097$\pm$0.008} & \textbf{0.029$\pm$0.005} \\
    \textbf{LSTM} &            $\Delta t = 0.025$ &     (high) &          0.050$\pm$0.001 &          0.163$\pm$0.001 &          0.082$\pm$0.001 \\

\hline 

\multicolumn{6}{l}{$N_H = 1000$, $N_L = 4000$} \\
\hline 

     \textbf{DON} &            $\Delta t = 0.025$ &     (high) &          0.123$\pm$0.010 &          0.234$\pm$0.006 &          0.168$\pm$0.009 \\
     \textbf{DON} &            $\Delta t = 0.125$ &      (low) &          0.047$\pm$0.006 &          0.116$\pm$0.004 &          0.041$\pm$0.003 \\
     \textbf{DON} & $\Delta t = \{0.025, 0.125\}$ &    (multi) &          0.048$\pm$0.002 &          0.121$\pm$0.007 &          0.045$\pm$0.005 \\
\textbf{DON-LSTM} &            $\Delta t = 0.025$ &     (high) &          0.024$\pm$0.004 &          0.053$\pm$0.008 &          0.009$\pm$0.003 \\
\textbf{DON-LSTM} & $\Delta t = \{0.025, 0.125\}$ &    (multi) & \textbf{0.023$\pm$0.001} & \textbf{0.096$\pm$0.009} & \textbf{0.028$\pm$0.005} \\
    \textbf{LSTM} &            $\Delta t = 0.025$ &     (high) &          0.039$\pm$0.003 &          0.135$\pm$0.004 &          0.056$\pm$0.003 \\

\hline 

\end{tabular}
\caption{The mean and the standard deviation of prediction metrics on high-resolution data. Each model is trained 5 times.}
\end{table}

\begin{table}[ht]
  \centering
\begin{tabular}{lllccc}
\multicolumn{6}{c}{\textbf{Viscous Burgers' equation ($\nu = 0.001$)}} \\\hline 

\textbf{Model} & \multicolumn{2}{l}{\textbf{Resolution}}  &              \textbf{MAE} &             \textbf{RMSE} &              \textbf{RSE} \\
\hline

\multicolumn{6}{l}{$N_H = 100$, $N_L = 400$} \\
\hline 

     \textbf{DON} &           $\Delta t = 0.01$ &     (high) &          0.139$\pm$0.001 &          0.207$\pm$0.002 &          0.104$\pm$0.002 \\
     \textbf{DON} &           $\Delta t = 0.05$ &      (low) &          0.109$\pm$0.001 &          0.161$\pm$0.002 &          0.063$\pm$0.002 \\
     \textbf{DON} & $\Delta t = \{0.01, 0.05\}$ &    (multi) &          0.108$\pm$0.001 &          0.158$\pm$0.002 &          0.070$\pm$0.002 \\
    \textbf{DON-LSTM} &           $\Delta t = 0.01$ &     (high) &          0.157$\pm$0.001 &          0.233$\pm$0.002 &          0.132$\pm$0.003 \\
    \textbf{DON-LSTM} & $\Delta t = \{0.01, 0.05\}$ &    (multi) & \textbf{0.076$\pm$0.005} & \textbf{0.131$\pm$0.007} & \textbf{0.042$\pm$0.004} \\
    \textbf{LSTM} &           $\Delta t = 0.01$ &     (high) &          0.106$\pm$0.003 &          0.169$\pm$0.002 &          0.069$\pm$0.002 \\
 
\hline 

\multicolumn{6}{l}{$N_H = 250$, $N_L = 1000$} \\
\hline 

     \textbf{DON} &           $\Delta t = 0.01$ &     (high) &          0.120$\pm$0.002 &          0.177$\pm$0.002 &          0.076$\pm$0.002 \\
     \textbf{DON} &           $\Delta t = 0.05$ &      (low) &          0.094$\pm$0.001 &          0.140$\pm$0.001 &          0.048$\pm$0.001 \\
     \textbf{DON} & $\Delta t = \{0.01, 0.05\}$ &    (multi) &          0.092$\pm$0.001 &          0.138$\pm$0.001 &          0.048$\pm$0.001 \\
\textbf{DON-LSTM} &           $\Delta t = 0.01$ &     (high) &          0.132$\pm$0.002 &          0.208$\pm$0.004 &          0.106$\pm$0.004 \\
\textbf{DON-LSTM} & $\Delta t = \{0.01, 0.05\}$ &    (multi) & \textbf{0.055$\pm$0.003} & \textbf{0.103$\pm$0.003} & \textbf{0.026$\pm$0.001} \\
    \textbf{LSTM} &           $\Delta t = 0.01$ &     (high) &          0.062$\pm$0.002 &          0.116$\pm$0.001 &          0.033$\pm$0.000 \\

\hline 
\multicolumn{6}{l}{$N_H = 400$, $N_L = 1600$} \\
\hline 

     \textbf{DON} &           $\Delta t = 0.01$ &     (high) &          0.109$\pm$0.001 &          0.160$\pm$0.001 &          0.062$\pm$0.001 \\
     \textbf{DON} &           $\Delta t = 0.05$ &      (low) &          0.085$\pm$0.001 &          0.126$\pm$0.001 &          0.039$\pm$0.001 \\
     \textbf{DON} & $\Delta t = \{0.01, 0.05\}$ &    (multi) &          0.083$\pm$0.001 &          0.124$\pm$0.001 &          0.040$\pm$0.001 \\
\textbf{DON-LSTM} &           $\Delta t = 0.01$ &     (high) &          0.108$\pm$0.003 &          0.187$\pm$0.004 &          0.086$\pm$0.004 \\
\textbf{DON-LSTM} & $\Delta t = \{0.01, 0.05\}$ &    (multi) & \textbf{0.044$\pm$0.001} & \textbf{0.085$\pm$0.001} & \textbf{0.017$\pm$0.001} \\
    \textbf{LSTM} &           $\Delta t = 0.01$ &     (high) &          0.050$\pm$0.001 &          0.098$\pm$0.001 &          0.023$\pm$0.001 \\
 
\hline 

\multicolumn{6}{l}{$N_H = 550$, $N_L = 2200$} \\
\hline 

     \textbf{DON} &           $\Delta t = 0.01$ &     (high) &          0.104$\pm$0.001 &          0.152$\pm$0.002 &          0.056$\pm$0.001 \\
     \textbf{DON} &           $\Delta t = 0.05$ &      (low) &          0.077$\pm$0.001 &          0.116$\pm$0.001 &          0.033$\pm$0.000 \\
     \textbf{DON} & $\Delta t = \{0.01, 0.05\}$ &    (multi) &          0.075$\pm$0.001 &          0.114$\pm$0.001 &          0.033$\pm$0.001 \\
\textbf{DON-LSTM} &           $\Delta t = 0.01$ &     (high) &          0.087$\pm$0.004 &          0.161$\pm$0.007 &          0.063$\pm$0.005 \\
\textbf{DON-LSTM} & $\Delta t = \{0.01, 0.05\}$ &    (multi) & \textbf{0.037$\pm$0.001} & \textbf{0.074$\pm$0.001} & \textbf{0.013$\pm$0.000} \\
    \textbf{LSTM} &           $\Delta t = 0.01$ &     (high) &          0.042$\pm$0.002 &          0.087$\pm$0.001 &          0.019$\pm$0.001 \\

\hline 
\multicolumn{6}{l}{$N_H = 700$, $N_L = 2800$} \\
\hline 

     \textbf{DON} &           $\Delta t = 0.01$ &     (high) &          0.099$\pm$0.001 &          0.146$\pm$0.002 &          0.052$\pm$0.002 \\
     \textbf{DON} &           $\Delta t = 0.05$ &      (low) &          0.078$\pm$0.001 &          0.118$\pm$0.001 &          0.034$\pm$0.001 \\
     \textbf{DON} & $\Delta t = \{0.01, 0.05\}$ &    (multi) &          0.075$\pm$0.000 &          0.113$\pm$0.001 &          0.032$\pm$0.000 \\
\textbf{DON-LSTM} &           $\Delta t = 0.01$ &     (high) &          0.071$\pm$0.003 &          0.140$\pm$0.006 &          0.048$\pm$0.004 \\
\textbf{DON-LSTM} & $\Delta t = \{0.01, 0.05\}$ &    (multi) & \textbf{0.033$\pm$0.000} & \textbf{0.068$\pm$0.001} & \textbf{0.011$\pm$0.000} \\
    \textbf{LSTM} &           $\Delta t = 0.01$ &     (high) &          0.036$\pm$0.001 &          0.079$\pm$0.001 &          0.015$\pm$0.000 \\
 
\hline 

\end{tabular}
\caption{The mean and the standard deviation of prediction metrics on high-resolution data. Each model is trained 5 times.}
\end{table}

\begin{table}[ht]
  \centering
\begin{tabular}{lllccc}
 
\multicolumn{6}{c}{\textbf{Benjamin--Bona--Mahony}} \\
\hline 
\textbf{Model} & \multicolumn{2}{l}{\textbf{Resolution}}  &              \textbf{MAE} &             \textbf{RMSE} &              \textbf{RSE} \\
\hline 

\multicolumn{6}{l}{$N_H = 250$, $N_L = 1000$} \\
\hline 
     \textbf{DON} &            $\Delta t = 0.075$ &     (high) &          0.554$\pm$0.019 &          0.932$\pm$0.031 &          0.297$\pm$0.020 \\
     \textbf{DON} &            $\Delta t = 0.375$ &      (low) &          0.166$\pm$0.012 &          0.383$\pm$0.011 &          0.050$\pm$0.003 \\
     \textbf{DON} & $\Delta t = \{0.075, 0.375\}$ &    (multi) &          0.155$\pm$0.011 &          0.331$\pm$0.011 &          0.041$\pm$0.003 \\
\textbf{DON-LSTM} &            $\Delta t = 0.075$ &     (high) &          0.198$\pm$0.041 &          0.510$\pm$0.095 &          0.092$\pm$0.034 \\
\textbf{DON-LSTM} & $\Delta t = \{0.075, 0.375\}$ &    (multi) & \textbf{0.088$\pm$0.005} & \textbf{0.271$\pm$0.036} & \textbf{0.025$\pm$0.007} \\
    \textbf{LSTM} &            $\Delta t = 0.075$ &     (high) &          0.191$\pm$0.008 &          0.544$\pm$0.013 &          0.101$\pm$0.005 \\

\hline 

\multicolumn{6}{l}{$N_H = 500$, $N_L = 2000$} \\
\hline 
     \textbf{DON} &            $\Delta t = 0.075$ &     (high) &          0.250$\pm$0.010 &          0.530$\pm$0.011 &          0.096$\pm$0.004 \\
     \textbf{DON} &            $\Delta t = 0.375$ &      (low) &          0.096$\pm$0.006 &          0.245$\pm$0.010 &          0.021$\pm$0.002 \\
     \textbf{DON} & $\Delta t = \{0.075, 0.375\}$ &    (multi) &          0.074$\pm$0.008 &          0.192$\pm$0.010 &          0.013$\pm$0.001 \\
\textbf{DON-LSTM} &            $\Delta t = 0.075$ &     (high) &          0.107$\pm$0.020 &          0.264$\pm$0.033 &          0.024$\pm$0.006 \\
\textbf{DON-LSTM} & $\Delta t = \{0.075, 0.375\}$ &    (multi) & \textbf{0.042$\pm$0.002} & \textbf{0.135$\pm$0.020} & \textbf{0.006$\pm$0.002} \\
    \textbf{LSTM} &            $\Delta t = 0.075$ &     (high) &          0.111$\pm$0.018 &          0.344$\pm$0.014 &          0.041$\pm$0.003 \\

\hline 
\multicolumn{6}{l}{$N_H = 750$, $N_L = 3000$} \\
\hline 
     \textbf{DON} &            $\Delta t = 0.075$ &     (high) &          0.181$\pm$0.018 &          0.380$\pm$0.008 &          0.050$\pm$0.002 \\
     \textbf{DON} &            $\Delta t = 0.375$ &      (low) &          0.058$\pm$0.006 &          0.168$\pm$0.005 &          0.010$\pm$0.001 \\
     \textbf{DON} & $\Delta t = \{0.075, 0.375\}$ &    (multi) &          0.046$\pm$0.002 &          0.140$\pm$0.012 &          0.007$\pm$0.001 \\
\textbf{DON-LSTM} &            $\Delta t = 0.075$ &     (high) &          0.048$\pm$0.007 &          0.119$\pm$0.025 &          0.005$\pm$0.002 \\
\textbf{DON-LSTM} & $\Delta t = \{0.075, 0.375\}$ &    (multi) & \textbf{0.032$\pm$0.006} & \textbf{0.125$\pm$0.022} & \textbf{0.005$\pm$0.002} \\
    \textbf{LSTM} &            $\Delta t = 0.075$ &     (high) &          0.068$\pm$0.005 &          0.250$\pm$0.008 &          0.021$\pm$0.001 \\

\hline 
\multicolumn{6}{l}{$N_H = 1000$, $N_L = 4000$} \\
\hline 
     \textbf{DON} &            $\Delta t = 0.075$ &     (high) &          0.125$\pm$0.009 &          0.288$\pm$0.012 &          0.028$\pm$0.002 \\
     \textbf{DON} &            $\Delta t = 0.375$ &      (low) &          0.043$\pm$0.002 &          0.110$\pm$0.001 &          0.004$\pm$0.000 \\
     \textbf{DON} & $\Delta t = \{0.075, 0.375\}$ &    (multi) &          0.033$\pm$0.006 &          0.103$\pm$0.011 &          0.004$\pm$0.001 \\
\textbf{DON-LSTM} &            $\Delta t = 0.075$ &     (high) &          0.061$\pm$0.023 &          0.162$\pm$0.071 &          0.011$\pm$0.008 \\
\textbf{DON-LSTM} & $\Delta t = \{0.075, 0.375\}$ &    (multi) & \textbf{0.019$\pm$0.001} & \textbf{0.074$\pm$0.011} & \textbf{0.002$\pm$0.001} \\
    \textbf{LSTM} &            $\Delta t = 0.075$ &     (high) &          0.056$\pm$0.008 &          0.191$\pm$0.013 &          0.012$\pm$0.002 \\

\hline 
\end{tabular}
\caption{The mean and the standard deviation of prediction metrics on high-resolution data. Each model is trained 5 times.}
\end{table}

\begin{table}[ht]
  \centering
\begin{tabular}{lllccc} 

\multicolumn{6}{c}{\textbf{Cahn--Hilliard equation}} \\\hline 

\textbf{Model} & \multicolumn{2}{l}{\textbf{Resolution}}  &              \textbf{MAE} &             \textbf{RMSE} &              \textbf{RSE} \\
\hline

\multicolumn{6}{l}{$N_H = 250$, $N_L = 1000$} \\
\hline 

     \textbf{DON} &          $\Delta t = 0.02$ &     (high) &          0.066$\pm$0.004 &          0.104$\pm$0.005 &          0.039$\pm$0.004 \\
     \textbf{DON} &           $\Delta t = 0.1$ &      (low) &          0.050$\pm$0.006 &          0.074$\pm$0.007 &          0.020$\pm$0.004 \\
     \textbf{DON} & $\Delta t = \{0.02, 0.1\}$ &    (multi) &          0.031$\pm$0.003 &          0.053$\pm$0.004 &          0.010$\pm$0.002 \\
\textbf{DON-LSTM} &          $\Delta t = 0.02$ &     (high) &          0.125$\pm$0.015 &          0.206$\pm$0.021 &          0.154$\pm$0.032 \\
\textbf{DON-LSTM} & $\Delta t = \{0.02, 0.1\}$ &    (multi) & \textbf{0.023$\pm$0.003} & \textbf{0.043$\pm$0.004} & \textbf{0.007$\pm$0.001} \\
    \textbf{LSTM} &          $\Delta t = 0.02$ &     (high) &          0.025$\pm$0.000 &          0.054$\pm$0.001 &          0.011$\pm$0.000 \\

\hline 

\multicolumn{6}{l}{$N_H = 500$, $N_L = 2000$} \\
\hline 

     \textbf{DON} &          $\Delta t = 0.02$ &     (high) &          0.041$\pm$0.001 &          0.067$\pm$0.001 &          0.016$\pm$0.001 \\
     \textbf{DON} &           $\Delta t = 0.1$ &      (low) &          0.044$\pm$0.002 &          0.063$\pm$0.004 &          0.014$\pm$0.002 \\
     \textbf{DON} & $\Delta t = \{0.02, 0.1\}$ &    (multi) &          0.018$\pm$0.001 &          0.031$\pm$0.002 &          0.003$\pm$0.000 \\
\textbf{DON-LSTM} &          $\Delta t = 0.02$ &     (high) &          0.079$\pm$0.010 &          0.141$\pm$0.013 &          0.072$\pm$0.014 \\
\textbf{DON-LSTM} & $\Delta t = \{0.02, 0.1\}$ &    (multi) & \textbf{0.016$\pm$0.002} & \textbf{0.030$\pm$0.003} & \textbf{0.003$\pm$0.001} \\
    \textbf{LSTM} &          $\Delta t = 0.02$ &     (high) &          0.015$\pm$0.000 &          0.037$\pm$0.001 &          0.005$\pm$0.000 \\

\hline 

\multicolumn{6}{l}{$N_H = 750$, $N_L = 3000$} \\
\hline 

     \textbf{DON} &          $\Delta t = 0.02$ &     (high) &          0.034$\pm$0.002 &          0.056$\pm$0.002 &          0.011$\pm$0.001 \\
     \textbf{DON} &           $\Delta t = 0.1$ &      (low) &          0.047$\pm$0.004 &          0.066$\pm$0.006 &          0.016$\pm$0.003 \\
     \textbf{DON} & $\Delta t = \{0.02, 0.1\}$ &    (multi) &          0.013$\pm$0.001 &          0.024$\pm$0.001 &          0.002$\pm$0.000 \\
\textbf{DON-LSTM} &          $\Delta t = 0.02$ &     (high) &          0.055$\pm$0.007 &          0.109$\pm$0.013 &          0.043$\pm$0.010 \\
\textbf{DON-LSTM} & $\Delta t = \{0.02, 0.1\}$ &    (multi) & \textbf{0.011$\pm$0.001} & \textbf{0.021$\pm$0.002} & \textbf{0.002$\pm$0.000} \\
    \textbf{LSTM} &          $\Delta t = 0.02$ &     (high) &          0.013$\pm$0.001 &          0.029$\pm$0.001 &          0.003$\pm$0.000 \\

\hline 

\multicolumn{6}{l}{$N_H = 1000$, $N_L = 4000$} \\

\hline 
     \textbf{DON} &          $\Delta t = 0.02$ &     (high) &          0.025$\pm$0.002 &          0.044$\pm$0.001 &          0.007$\pm$0.000 \\
     \textbf{DON} &           $\Delta t = 0.1$ &      (low) &          0.010$\pm$0.000 &          0.020$\pm$0.001 &          0.001$\pm$0.000 \\
     \textbf{DON} & $\Delta t = \{0.02, 0.1\}$ &    (multi) &          0.009$\pm$0.000 &          0.016$\pm$0.000 &          0.001$\pm$0.000 \\
\textbf{DON-LSTM} &          $\Delta t = 0.02$ &     (high) &          0.045$\pm$0.003 &          0.092$\pm$0.007 &          0.031$\pm$0.004 \\
\textbf{DON-LSTM} & $\Delta t = \{0.02, 0.1\}$ &    (multi) & \textbf{0.007$\pm$0.000} & \textbf{0.015$\pm$0.000} & \textbf{0.001$\pm$0.000} \\
    \textbf{LSTM} &          $\Delta t = 0.02$ &     (high) &          0.011$\pm$0.001 &          0.025$\pm$0.001 &          0.002$\pm$0.000 \\

\hline 
\end{tabular}
\caption{The mean and the standard deviation of prediction metrics on high-resolution data. Each model is trained 5 times.}
\end{table}
\FloatBarrier

\section{Architecture details}

\label{app:architectures} 

\subsection{DeepONet}

The vanilla DeepONet is constructed with a branch and trunk network with the same output sizes, which are merged using Einstein summation (\autoref{fig:proposed_architecture}). The input to the branch network is of the size [batch\_size, x\_len], the input to the trunk network is of the size [x\_len$\times$t\_len, 2], and the output of the network after the Einstein summation is [batch\_size, x\_len$\times$t\_len], where $\text{batch\_size}$ refers to the number of samples in each minibatch, and x\_len and t\_len, to the sizes of the spatial and temporal dimensions, \textit{i.e.}, the number of discretization points. 

\begin{table}[h]
\caption{Branch network.}
\label{don_branch_architecture}
\vskip 0.15in
\begin{center}
\begin{small}
\begin{sc}
\begin{tabular}{llccr}
 & Layer & Output shape & Activation & Param   \\
\hline
0   & Input &   (None, x) & - & $0$ \\
1   & Dense &   (None, 150) & swish & $150x+150$ \\
2   & Dense &   (None, 250)  & swish & $37,750$ \\
3   & Dense &   (None, 450)  & swish & $112,950$ \\
4   & Dense &   (None, 380)  & swish & $171,380$ \\
5   & Dense &   (None, 320)  & swish & $121,920$ \\
6   & Dense &   (None, 300)  & linear & $96,300$ \\

\hline
\end{tabular}
\end{sc}
\end{small}
\end{center}
\vskip -0.1in
\end{table}

\begin{table}[h]
\caption{Trunk network.}% Total nr of parameters is $380,750$.}
\label{don_trunk_architecture}
\vskip 0.15in
\begin{center}
\begin{small}
\begin{sc}
\begin{tabular}{llccr}
 & Layer & Output shape & Activation & Param   \\
\hline
0   & Input &   (None, 2) & - & 0 \\
1   & Dense &   (None, 200) & swish & $600$ \\
2   & Dense &   (None, 220)  & swish & $44,220$ \\
3   & Dense &   (None, 240)  & swish & $53,040$ \\
4   & Dense &   (None, 250)  & swish & $60,250$ \\
5   & Dense &   (None, 260)  & swish & $65,260$ \\
6   & Dense &   (None, 280)  & linear & $73,080$ \\
7   & Dense &   (None, 300)  & linear & $84,300$ \\

\hline
\end{tabular}
\end{sc}
\end{small}
\end{center}
\vskip -0.1in
\end{table}

%The layers of the branch net are gradually wider and then narrower, the layers of the trunk net are widening up. 

\FloatBarrier

The DeepONet is trained in minibatches of $50$ samples using the Adam optimizer and learning rate of $1e-4$. For the DeepONet the data is normalized in the following manner: the inputs to the branch network and the outputs of the DeepONet use standard scaling, and the inputs to the trunk network use min-max (\autoref{appendix:normalization}). The vanilla DeepONet for the full trajectory is trained up to $25,000$ epochs.

\subsection{DON-LSTM}

\begin{table}[h]
\caption{LSTM extension for the input of $t$ timesteps.} %Total nr of parameters is $60,250$.}
\label{rnn_extension_architecture}
\vskip 0.15in
\begin{center}
\begin{small}
\begin{sc}
\begin{tabular}{llccr}
 & Layer & Output shape & Activation & Param   \\
\hline
0   & DeepONet output &   (None, $x \times t$) & - & $0$ \\
1   & Reshape &   (None, $t$, $x$) & - &  $0$\\
2   & LSTM &   (None, $t$, 200)  & tanh &  $4\times ((x+1)\times200+200^2)$ \\
3   & Dense &   (None, $t$, $x$)  & linear & $xt+x$ \\

\hline
\end{tabular}
\end{sc}
\end{small}
\end{center}
\vskip -0.1in
\end{table}

\FloatBarrier

The inputs and outputs in the LSTM training are normalized using standard scaling.  

%When using gated units instead of the simple RNN, the number of parameters is considerably higher. For GRU, Layer \#2 has $151,200$ parameters and the whole network has $161,250$ parameters. For LSTM, Layer \#2 has $200,800$ parameters and the whole network has $210,850$ parameters.

\FloatBarrier

\section{Data scaling details}
\label{appendix:normalization}

When training the DeepONets, the inputs to the branch net are scaled using standard scaling, while the inputs to the trunk net use the min-max scaling. For the LSTM, the inputs are scaled using standard scaling. The outputs are always scaled with the standard scaling.

\subsection{Standard scaling}

The standard scaling formula is defined as:
\begin{equation}
    x' = \frac{x - \mu}{\sigma},    
\end{equation}

where $x'$ are the standardized values, $x$ are the original values, $\mu$ is the mean, and $\sigma$ is the standard deviation of $x$.

\subsection{Min-max scaling}

Min-max scaling, also known as Min-max normalization, scales the data between 0 and 1. It is expressed by the equation:

\begin{equation}
x' = \frac{x - x_{min}}{x_{max} - x_{min}},
\end{equation}

where $x'$ are the normalized values, $x$ are the original values, and $x_{min}$ and $x_{max}$ are the minimal and maximal values of $x$.

\section{Evaluation metrics}

\label{app:eval_metrics_explained}

In this study, we used several metrics to evaluate the performance of the model such as mean average error, root mean squared error, and relative squared error.

\subsection{Mean average error}

The mean average error (MAE) is expressed as:
\begin{equation}
    \text{MAE} = \frac{1}{n} \sum_{i=1}^{n} | y_i - \hat{y_i} |,
\end{equation}

where $n$ is the number of samples, $y_i$ is the true value of the $i^{th}$ sample, and $\hat{y_i}$ is the predicted value of the $i^{th}$ sample.

\subsection{Root mean squared error}

The root mean squared error (RMSE) is expressed as:
\begin{equation}    
    \text{RMSE} = \sqrt{\frac{1}{n}\sum_{i=1}^{n}(y_i - \hat{y_i})^2},
\end{equation}

where: $n$ is the number of samples, $y_i$ is the true value of the $i^{th}$ sample and $\hat{y_i}$ is the predicted value of the $i^{th}$ sample.

\subsection{Relative squared error}

\label{app:rse}

The relative squared error (RSE) is the total squared error between the predicted values and the ground truth normalized by the total squared error between the ground truth and the mean. The metric is expressed as:

\begin{equation}
    \text{RSE} = 
    \frac{
        \frac{1}{n}\sum_{i=1}^{n}(y_i - \hat{y_i})^2
        }
        {
        \frac{1}{n} \sum_{i=1}^{n}(y_i - \bar{y})^2
        }
\end{equation}

where $y_i$ is the true value of the $i^{th}$ sample, $\hat{y_i}$ is the predicted value of the $i^{th}$ sample, and $\bar{y}$ is the mean value of all samples.

\end{document}

%% file: math_commands.tex
\usepackage{amsmath,amsfonts,bm}

\def\eqref#1{equation~\ref{#1}}
\def\1{\bm{1}}

\DeclareMathAlphabet{\mathsfit}{\encodingdefault}{\sfdefault}{m}{sl}
\SetMathAlphabet{\mathsfit}{bold}{\encodingdefault}{\sfdefault}{bx}{n}